\documentclass{clv3}
\usepackage{times}
\usepackage{tikz} 
\usepackage{latexsym}

\usepackage{url}

\usepackage{hyperref}
\usepackage{xcolor}
\definecolor{darkblue}{rgb}{0, 0, 0.5}
\hypersetup{colorlinks=true,citecolor=darkblue, linkcolor=darkblue, urlcolor=darkblue}

\issue{1}{1}{2016}

\dochead{}

\runningtitle{On Learning Interpreted Languages with Recurrent Models}

\runningauthor{Denis Paperno}

\begin{document}

\title{On Learning Interpreted Languages with Recurrent Models}

\author{Denis Paperno
  Utrecht University\\
Trans 10, 3512 JK Utrecht \\
\texttt{d.paperno@uu.nl} \\
  }


\maketitle

\begin{abstract}
Can recurrent neural nets, inspired by human sequential data processing, learn to understand language? We construct simplified datasets reflecting core properties of natural language as modeled in formal syntax and semantics: recursive syntactic structure and compositionality. We find LSTM and GRU networks to generalise to compositional interpretation well, but only in the most favorable learning settings, with a well-paced curriculum, extensive training data, and left-to-right (but not right-to-left) composition.
\end{abstract}

\section{Introduction}
Common concerns in NLP are that current models tend to be very data hungry and rely on shallow cues and biases in the data \citep[e.g.~][]{geva2019we}, failing at deeper, human-like generalization.  In this article, we present a new task of acquiring compositional generalization from a small number of linguistic examples. Our setup is inspired by properties of language understanding in humans.

First, natural language exploits recursion: natural language syntax consists of constructions, represented in formal grammars as rewrite rules, which can recursively embed other constructions of the same kind. 
For example, noun phrases can consist of a single proper noun (e.g. \textit{Ann}) but can also, among other possibilities, be built from other noun phrases recursively via the possessive construction, as in \textit{Ann's child}, \textit{Ann's child's friend}, \textit{Ann's child's friend's parent} etc.

Second, recursive syntactic structure drives semantic interpretation. The meaning of the noun phrase \textit{Ann's child's friend} is not the sum of the meanings of the individual words (otherwise it would have been equivalent to \textit{Ann's friend's child}). To interpret a complex expression, one has to follow its syntactic structure, first identifying the meaning of the subconstituent (\textit{Ann's friend}), and then computing the meaning of the whole. Semantic composition, argued to be a core component of language processing in the human brain \citep{mollica2020composition}, can be formalized as function application, where one constituent in a complex structure corresponds to an argument of a function that another constituent encodes. For instance, in \textit{Ann's child}, we can think of \textit{Ann} as denoting an individual and \textit{child} as denoting a function from individuals to individuals. In formal semantics, function argument application is the basic compositionality mechanism, extending to a wide range of constructions.
The focus on recursion in semantic composition differentiates our approach from other attempts at examining compositional properties of neural systems \citep{ettinger2018assessing,soulos2019discovering,
andreas2019measuring,
mickus2020meaning}; \citet{kim2020cogs} includes depth of recursion as one of the many aspects of systematic semantic generalization. The work on compositionality learning most closely related to ours  relies on artificial languages of arithmetic expressions and sequence operations \citep{Hupkes:2018:VCR:3241691.3241713,listops,hupkes2020compositionality}. 

Third, natural language interpretation, while being sensitive to syntactic structure, is robust to syntactic variation. For example, humans are equally capable of learning to interpret and using left-branching structures such as $\texttt{NP} \longrightarrow \texttt{NP} \textit{'s} \texttt{ N}$ (\textit{Ann's child}) and right-branching structures such as $\texttt{NP} \longrightarrow \textit{the \texttt{N } \textit{of} \texttt{ NP}}$ (\textit{the child of Ann}).
Finally, language processing in humans is  sequential; people process and interpret linguistic input on the fly, without much lookahead or waiting for the linguistic structure to be completed. 
This property has diverse consequences for language and cognition \citep{christiansen2016now}, and gives a certain degree of cognitive plausibility to \textit{unidirectional recurrent models } compared to other neural architectures. 

To summarize, in order to mimic human language capacities, an artificial system must be able to 1) learn languages with compositionally interpreted recursive structures, 2) adapt to surface variation in the syntactic patterns, and 3) process its inputs sequentially. Recurrent neural networks are promising with respect to these desiderata; in a syntactic task \citep{LAKRETZ2021104699}, they demonstrated similar error patterns to humans in processing recursive structures. We proceed now to defining a semantic task that allows us to directly test models on these capacities.

\section{The task}\label{task}

Our approach builds closely on natural language and systematically explores the factor of syntactic structure (right vs.~left branching). We define toy interpreted languages based on a fragment of English. 
Data was generated with the following grammars:

Left-branching language: $\textsc{NP}\longrightarrow \textsc{Name}$, $\textsc{NP}\longrightarrow \textsc{NP}\texttt{ 's }\textsc{RN}$

Right-branching language: $\textsc{NP}\longrightarrow \textsc{Name}$, $\textsc{NP}\longrightarrow \texttt{the }\textsc{RN}\texttt{ of }\textsc{NP}$

Both: \textsc{Name}: \texttt{Ann}, \texttt{Bill}, \texttt{Garry}, \texttt{Donna}; \textsc{RN}: \texttt{child}, \texttt{parent}, \texttt{friend}, \texttt{enemy}


\begin{figure}
\begin{center}
\begin{tikzpicture}
\path
(0,0cm) node (b0) {$\texttt{\small Ann}$} 
+ (1,0) node (b1) {$\texttt{\small Donna}$}
+ (2,0) node (b2) {$\texttt{\small Garry}$}
+ (3,0) node (b3) {$\texttt{\small Bill}$}
+ (0.63,0.47) node  {$\textrm{\tiny parent}$}
+ (1.6,0.47) node  {$\textrm{\tiny parent}$}
+ (1.6,1.05) node  {$\textrm{\tiny parent}$}
+ (1.6,-1.05) node  {$\textrm{\tiny enemies}$}
+ (0.63,-0.49) node  {$\textrm{\tiny friends}$}
+ (1.6,-0.49) node  {$\textrm{\tiny enemies}$}
+ (2.4,0.47) node  {$\textrm{\tiny parent}$}
+ (2.48,-0.49) node  {$\textrm{\tiny friends}$};
\draw[<-] (3.14,0.27) arc (30:150:1.8cm);
\draw[<->] (3.14,-0.27) arc (-30:-150:1.8cm);
\draw[->] (1,0.27) arc (60:120:0.9cm);
\draw[->] (2,0.27) arc (60:120:0.9cm);
\draw[<->] (1,-0.27) arc (-60:-120:0.9cm);
\draw[<->] (2,-0.27) arc (-60:-120:0.9cm);
\draw[->] (3,0.27) arc (60:120:0.9cm);
\draw[<->] (3,-0.27) arc (-60:-120:0.9cm);
\end{tikzpicture}
\end{center}
\caption{\label{friend-figure} An example of a toy universe with relations indicated by arrows. The \textit{child} relation is the inverse of \textit{parent}.}
\end{figure}

%

\begin{table}[t!]
\begin{center}
\begin{tabular}{|r|lcc|}
\hline example & label & complexity & partition\\ \hline
Garry & Garry & 1 & train \\
Bill's enemy & Ann & 2 & train\\
Ann's friend & Donna & 2 & train\\
Donna's parent & Garry & 2 & train\\
Ann's child & Donna & 2 & train\\
Garry's enemy's parent & Ann & 3 & train\\
\hline
Ann's friend's parent & Garry & 3 & validation\\
\hline
Bill's enemy's child & Donna & 3 & test\\
\hline
\end{tabular}
\end{center}
\caption{\label{gen-examples} Some examples from a dataset with top complexity 3 illustrating the targeted semantic generalization.}
\end{table}

Interpretation is defined model-theoretically. We randomly generate a model where each proper name corresponds to a distinct individual and each function denoted by a common noun is total, e.g.~every individual has exactly one enemy, exactly one friend, etc. This total function interpretation simplifies natural language patterns but is an appropriate idealization for our purposes. An example a model with all relations is given in Figure~\ref{friend-figure}. 

In such a model, each well-formed expression of the language is an individual identifier. The denotation of any expression regardless of its complexity (= number of content words) can be calculated by recursive application of functions to arguments, guided by the syntactic structure.

The task given to the models is to identify the individual that corresponds to each expression; e.g.~\textit{Ann's child's enemy} (complexity 3) is the same person as \textit{Bill} (complexity 1). Since there is just a finite number of individuals in any given model, the task formally boils down  to string classification, assigning each expression to one of the set of individuals in the model.

By its nature, solving the task presumes (implicitly) mastering parsing and semantic composition at the same time. Parsing is deciding what pieces of information from different points in the input string to combine in what order, for example in \textit{the father of the friend of Garry} one starts by combining \textit{friend} and \textit{Garry}, then the result with \textit{father}. Semantic composition is performing the actual combination, in this case by recursively applying functions to arguments: \texttt{father(friend(garry))}. Because of how the relations in the universe are randomly generated every time, there is no consistent bias for the model to exploit. So without a compositional strategy, the value of a held out complex example cannot be reliably determined. Therefore, if a system maps unseen complex expressions (\textit{the father of the friend of Garry}) to their referents with high accuracy, it must have approximated a compositional solution of the task.

\section{Datasets}
We use languages with four individuals and four relation nouns. The interpretations of the relation nouns were generated randomly while making sure that `friend' and `enemy' are symmetric and `parent' and `child' are antisymmetric and inverse of each other.  
The interpretation of the four function elements was randomly assigned for model initialization. We used all expressions of the language up to complexity $n$ as data; validation and testing data was randomly selected among examples of the maximal complexity $n$. Expressions of complexity 1 and 2 were always included in the training partition since they are necessary to learn the interpretation of lexical items. For example, the simplest set of training data (up to complexity 3) contained all names (examples of complexity 1), required to learn the individuals; all expressions with 2 content words like \textit{Ann's child}, necessary for learning the meanings of functional words like \textit{child}; and a random subset of three content word expressions like \textit{Ann's child's friend}, which might guide the systems to learning recursion. To further simplify the task, each dataset contained either only left branching or only right branching examples in all three partitions (training/validation/testing).
Dataset sizes are given in Table \ref{data-size}, which reports the default setup with 80\% examples of maximum complexity included in training data and the rest distributed between validation and test data. In this setup, the proportions of data partitions are approximately 85\% training, 8\% validation and 7\% testing.  

\begin{table}[t!]
\begin{center}
\begin{tabular}{|l|ccccc|}
\hline \bf test ex.~complexity: & \bf 3 & \bf 4 & \bf 5& \bf 6& \bf 7\\ \hline
train & 71 & 288  & 1159 & 4640 & 18567\\
dev & 7 & 28 & 112 & 451 & 1802\\
test & 6 & 24 & 93 & 369 & 1475\\
\hline
total & 84 & 340 & 1364 & 5460 & 21844\\
\hline
\end{tabular}
\end{center}
\caption{\label{data-size} Dataset sizes for each maximal complexity value. }
\end{table}

\section{Models and training}
We tested the learning capacities of standard recurrent neural models on our task: a vanilla recurrent neural network \citep[SRN,][]{elman1991distributed}, a long short-term memory network  \citep[LSTM,][]{hochreiter1997long}, and a gated recurrent unit \citep[GRU,][]{cho2014properties}.\footnote{Code used in the article is available at \url{https://github.com/dpaperno/LSTM\_composition}.}
 The systems were implemented in PyTorch and used a single hidden layer of 256 units. Using more layers or fewer units did not substantially improve, and sometimes dramatically deteriorated, the performance of the models. Models were trained from a randomly initialized state on tokenized input (e.g.\ \texttt{["Ann","s","child"]}). The training used Adam optimizer and negative log likelihood loss for sequence classification.
Each model was run repeatedly with different random initializations. Datasets were randomly generated for each initialization of each model.

 We also set a curriculum whereby the system is given training examples of minimal complexity first, with more complex examples added gradually in the process of training. The best curriculum settings varied by model and test example complexity, but on average, the most robust results were obtained by adding examples of the next complexity level after every ten epochs (\textit{gentle curriculum}). A curriculum very consistently led to an improvement over 
the alternative training whereby training examples of all complexity levels were available to the models at all epochs. For more details on alternative curricula, see the Appendix, section 1. 

\section{Results}

To treat complex expressions, the successful model needs to generalize by learning to compose the representations of simple expressions recursively. But representations of simple expressions (complexity 1 and 2) have to be memorized in one form or another because their interpretation is arbitrary;  without such memorization, generalization to complex inputs is impossible.


Like in related studies \citep[e.g.][]{DBLP:journals/corr/abs-1802-06467}, SRN was behind the gated architectures in accuracy, with a dramatic gap for longer examples. Accuracies of gated models are summarized in Table \ref{acc-table}; for more details, see Appendix, section 1. 
We find that gated architectures learn compositional interpretation in our task only in the best conditions. A curriculum proved essential for recursive generalization. Informally, the system has to learn to interpret words first, and recursive semantic composition has to be learned later. 

All recurrent models generalized only to left-branching structures; the accuracy in the right branching case is much lower. This means that the systems only learn to apply composition following the linear sequence of the input and fail when the order of composition as determined by the syntactic structure runs opposite to the linear order. In other words, the system manages to learn composition when the parse trivially corresponds to the default processing sequence, but fails when nontrivial parsing is involved. Therefore, semantic composition is mastered, but not simultaneously with the independent task of parsing; cf.\ discussion above in \ref{task}.


\begin{table}[t!]
\begin{center}
\begin{tabular}{|l|ccccc|}
\hline \bf test ex.~complexity: & \bf 3 & \bf 4 & \bf 5& \bf 6& \bf 7\\ \hline
right branching & 0.09 & 0.35 & 0.16 & 0.3 & 0.17\\
left branching &\bf 0.94&\bf 0.92&\bf 0.88&\bf 0.92 &\bf 0.92\\
left, no curriculum & 0.34 & 0.25 & 0.29 & 0.3 & 0.28\\
\hline
\end{tabular}
\end{center}
\caption{\label{acc-table} Accuracy of gated architectures as a function of the language and input data complexity. Results averaged over 10 random initializations of an LSTM and 10 random initializations of a GRU model. Training data in each run includes examples of complexity up to and including $n$, with testing data (disjoint from training) of complexity exactly $n$. Random baseline is 0.25. For more details, see Appendix, section 1.}
\end{table}

\section{Zero-shot Compositionality Testing}

How much recursive input does a successful system need to master recursive interpretation?  Ideally, learners with a strong bias towards languages with recursive syntactic structure (not an implausible assumption when it comes to human language learners) could acquire them in a zero-shot fashion, without any training examples that feature recursive structures. For example, assume that such a learner knows that the name \textit{Donna} and the phrase \textit{Ann's child} refer to the same individual,  and that \textit{Donna's enemy} refers to Bill. Without any exposure to examples longer than \textit{Donna's enemy}, the learner could still infer that  \textit{Ann's child's enemy} of unseen length 3 is also Bill. In a recurrent neural network, the expectation can be interpreted as follows: both \textit{Donna} and the phrase \textit{Ann's child} are expected to be mapped to similar hidden states, and since those hidden states allow the system to identify Bill after processing \textit{'s enemy} in \textit{Donna's enemy}, the same can be expected for the phrase \textit{Ann's child's enemy}.
To test whether such zero-shot (or few-shot) recursion capacity actually arises, we train the model while varying the amount of recursion examples available as training data from 0\% to 80\% of all examples of complexity 3. As shown in Table \ref{recurs-table}, the LSTM needs to be trained on a vast majority of recursive examples to be able to generalize. Similar results for GRU and SRN are reported in the Appendix, section 2.

\begin{table}[t!]
\begin{center}
\begin{tabular}{|l|ccccc|}
\hline share of recursive examples included in training & 0.0 & 0.2 & 0.4&  0.6 & 0.8\\ \hline training/testing examples & 20/29 & 32/24 & 45/18&  58/12 & 71/6\\ \hline
share of  training examples that are recursive & 0.0 & 0.375 & 0.556 & 0.655  & 0.718 \\ \hline
average accuracy & 0 & .61 & .76 & .89 & \bf .98\\
perfect accuracy & 0 & 0 & 0 & .4 & \bf .9\\
\hline
\end{tabular}
\end{center}
\caption{\label{recurs-table} LSTM's data hungriness in learning recursion. Percentage of complexity 3 data included in training vs.\ test accuracy. We report average accuracy over 10 runs (each with a random initialization of the model and data generation) as well as the share of runs with perfect accuracy.  Random baseline for accuracy is .25.}
\end{table}

Recursive compositionality does not come for free and has to be learned from extensive data. This observation goes in line with negative findings in literature on compositionality learning  \citep{DBLP:journals/corr/abs-1802-06467,DBLP:journals/corr/abs-1805-09657,lake2018generalization,NEURIPS2020_e5a90182}. 
Contrary to some of those findings, we do observe generalization to bigger structures \emph{after} substantial evidence for a compositional solution is available to the system. GRU and LSTM models trained on examples of complexities 1--3 perform well on examples of unseen greater lengths (96\% and 99\% accuracy on complexity 4, respectively; 95\% and 98\% on complexity 5 etc.). 
Only on complexity 7 did the trained LSTM show somewhat degraded performance (77\% accuracy), with GRU maintaining robust generalization (96\% accuracy). For detailed results, see Appendix, section 3. Interestingly, models trained on complexities 1--3 and generalizing recursively to unseen lengths sometimes performed better than the same models exposed to more complex inputs during training (Table 3). This further strengthens our conclusions above about the importance of curriculum: complex training examples might be detrimental for recursive generalization at late as well as early learning stages.


What drives generalization success in these settings? We hypothesize that the RNNs are good at approximating a finite state solution to the interpretation problem which could run as follows: when processing the token \textit{Ann}, an automaton would transition to the state `Ann'; after processing \textit{'s child}, they transition from `Ann' to the state `Donna'; etc. The state of the automaton after processing the whole sequence corresponds to the referent of the phrase. This particular finite state solution relies on left-to-right processing and only applies to left-branching structures (\textit{Ann's child's friend}) but not to right-branching ones (\textit{the friend of the child of Ann}).

\section{Discussion: the Left-Branching Asymmetry of Recurrent Models}

In our experiments, recurrent models generalize reliably only to left-branching structures. This suggests that RNNs, despite greater theoretical capacities, remain well-suited to sequential, left-to-right processing and do not learn a fully recursive solution to the interpretation problem. As literature suggests, recurrent models can generalize to non-left-branching data patterns if the language allows for a simpler interpretation strategy that minimizes computation over recursive structures, such as the `cumulative' strategy of \citet{Hupkes:2018:VCR:3241691.3241713}. However, a cumulative interpretation strategy might not be an easy solution to learn for natural language or artificial languages. Even in Hupkes et al., where a GRU model showed reasonable average performance for mixed-directionality structures, a much lower error level  is reported for left branching examples and a much higher error level for right branching examples.

The left-to-right branching asymmetry of recurrent models is doubtlessly related to their natural representational structure. When processing a left-branching structure left to right, at any point the system needs to `remember' only one element, the intermediate result of computation (an individual in our toy language, or a number in the arithmetic language). 
In the case of our language, the incremental interpretation of the left-branching model keeps track of the current individual (e.g.\ `Ann') and switches to the next individual according to the next relation term encountered (e.g.\ `friend'). For a universe with $k$ individuals, a sequential model must distinguish only $k$ states. If however one interprets right-branching examples (\textit{the father of the friend of Ann}), one has to keep a representation of a function to be applied to the individual whose name follows (example of such a function: \textit{the father of the friend}). There are $k^k$ $U\rightarrow U$ functions over the domain of individuals $U$, so the intermediate representation space must be significantly richer than in the left-branching case where only $k$ states suffice. The number of states is a problem not only from the viewpoint of model expressiveness, but also from the learning point of view: it is not realistic to observe all $k^k$ functions at training time. 

The brute force approach to left-to-right parsing or interpretation consists in keeping a stack of representations to be integrated at a later step; but in practice learned recurrent neural models are known to be too poor representationally to emulate stacks 
even if they are more powerful in practice than finite state machines \citep{weiss2018practical,bernardy2018can}. As Hupkes et al.'s analysis suggests, GRUs can emulate stack behavior to a restricted degree. The cumulative strategy that they argue the GRU learns makes use of stacks of binary values, but generalizing stack behavior to an unseen situation that required novel stack values apparently fails, as evidenced by the dramatically low performance on right branching examples of unseen complexity. Hupkes et al.'s models thus probably learn to operate only over small stacks by memorizing specific stack states and transitions between them (for their task, the relevant stack states are few in number). Their recurrent models can therefore be interpreted as finding ``approximate solutions by using
the stack as unstructured memory'' \cite{hao2018context}.

What solutions do recurrent models learn when they fail to generalize compositionally in the right branching case? To find out, we analysed the predictions of SRN, GRU and LSTM models for right branching languages, and found a consistent pattern. 
They do learn names correctly (e.g.~\textit{Ann}), but struggle when it comes to anything more complex. 
In a right branching input language, all models tend to `forget' long distance information and ignore anything but the suffix, assigning the same referent to \textit{the enemy of Ann}, \textit{the friend of the enemy of Ann}, \textit{the child of the enemy of Ann} etc. Often but not always models also overfit, predicting for complex examples the referent which occurred more often than others in the training data, even by a small margin. 
None of these observations pertain 
to learning left-branching patterns which are essentially a mirror image of right branching ones, 
although for them an RNN 
could just as easily fall back on suffix matching or the majority class. Instead, a converged model for left-branching data proceeds incrementally as per the expectations outlined above, correctly classifying \textit{Donna} as \textit{Donna}, \textit{Donna's parent} as \textit{Garry}, \textit{Donna's parent's friend} as \textit{Bill}, and so on for longer inputs. Crucial earlier parts of the string affect the output, and the majority class of the training data does not tend to come up as a prediction.

\section{Conclusion}
The results reported in this article both are encouraging and point to limitations of recurrent models. RNNs do learn compositional interpretation from data of small absolute size and generalize to more complex examples; certain favorable conditions are required, including a curriculum and a left branching language. Our findings are cautiously optimistic compared to claims in the literature about neural nets' qualitatively limited generalization capacity \citep{baroni2020linguistic}.
In the future, we plan 
to further explore the capacities of diverse models on our task. Indeed, other architectures might contain reasonable biases towards processing recursive languages not limited to the left branching case. 
Would it be beneficial, for instance, to augment the recurrent architecture with stack memory \citep{joulin2015inferring,yogatama2018memory}, add a chart parsing component \citep{le2015forest,maillard2017jointly}, or switch from sequential to attention-based processing altogether \citep{vaswani2017attention}? Or would this be possible only at the cost of the remarkable data efficiency we observed in recurrent models?

One idea seems especially appealing: if unidirectional RNN models can handle left recursive branching while right recursion is its mirror image, could bidirectoinal RNNs offer a general solution? There are however reasons for skepticism. Bidirectional models showed mixed results in our preliminary experiments, although further hyperparameter tuning and stricter regularization might lead to improvement. Further, even if biRNNs master both branching directions separately, they may still struggle with mixed structures which are common in natural language.
Lastly, whether human language learning has similar properties to those of RNNs presents an interesting area for exploration. Do humans, like our RNNs, require substantial data to learn recursion? 
And do humans, like our RNNs, have a bias towards left branching structures, which are easy for recursive sequential processing? (e.g.\ \textit{Garry's father}, rather than \textit{the father of Garry}). Suggestive evidence for the latter: left-branching possessive constructions emerge in infant speech even in languages that don't have them, e.g.\ \textit{Yael sefer} `Yael's book' (Hebrew, \citealt{armon1998mommy}) or 	\textit{zia trattore} `aunt's tractor' (Italian, \citealt{torregrossamelloni}).

\paragraph*{Acknowledgments} The research has been supported in part by CNRS PEPS ReSeRVe grant. I thank Germ\'{a}n Kruszewski, Marco Baroni, and anonymous reviewers for useful input.
\bibliography{emnlp2018}

\appendix
\section*{Appendices}
\label{sec:appendix}

\section{Recurrent architectures and alternative curricula}
LSTM, GRU, and Elman's Simple Recurrent Network (SRN) were implemented in Python 3.7.0 using built-in functions of PyTorch 1.5.0. Each model was trained for 100 epochs or until no improvement on the validation set was observed for 22 epochs. A number of curricula was compared. The default `gentle' curriculum consisted in adding examples of the next complexity level after every 10 epochs. No curriculum means that all training examples were used at all epochs. A slow curriculum consisted in adding examples of the next complexity level after every 20 epochs. A steep curriculum consisted in adding training examples of complexity 2 after 10 epochs and all other training examples after another 10 epochs. Results of the three architectures for all curricula are given in Tables \ref{acc-table-lstm}, \ref{acc-table-gru}, and \ref{acc-table-srn}.

\begin{table}[t!]
\begin{center}
\begin{tabular}{|l|ccccc|}
\hline \bf test ex.~complexity: & \bf 3 & \bf 4 & \bf 5& \bf 6& \bf 7\\ \hline
right branching & 0.02 & 0.29 & 0.12 & 0.23 & 0.1\\
left branching &0.9&\bf 0.9&\bf 0.76&\bf 0.93 &\bf 0.9\\
left, no curriculum & 0.23 & 0.2 & 0.24 & 0.26 & 0.25\\
left, steep curriculum & \bf 0.95 & 0.68 & 0.71 & 0.33 & 0.33 \\
left, slow curriculum & 0.42 & 0.73 & 0.55 & 0.69 & 0.61 \\
\hline
\end{tabular}
\end{center}
\caption{\label{acc-table-lstm} Accuracy of LSTM as a function of the language and input data complexity. Training data in each run includes examples of complexity up to and including $n$, testing data (disjoint from training) contained examples of complexity exactly $n$. Random baseline is 0.25.}
\end{table}

\begin{table}[t!]
\begin{center}
\begin{tabular}{|l|ccccc|}
\hline \bf test ex.~complexity: & \bf 3 & \bf 4 & \bf 5& \bf 6& \bf 7\\ \hline
right branching & 0.17 &  0.42 & 0.21 & 0.37 & 0.25\\
left branching &0.98 & 0.95 &\bf 1 & 0.92 & 0.94 \\
left, no curriculum & 0.45 & 0.3 & 0.34 & 0.34 & 0.31\\
left, steep curriculum & \bf 1 & 0.99 & 0.81 & 0.43 & 0.43 \\
left, slow curriculum & \bf 1 & \bf 1 & 0.997 & \bf 1 & \bf 0.997 \\
\hline
\end{tabular}
\end{center}
\caption{\label{acc-table-gru} Accuracy of GRU as a function of the language and input data complexity. Training data in each run includes examples of complexity up to and including $n$, testing data (disjoint from training) contained examples of complexity exactly $n$. Random baseline is 0.25.}
\end{table}

\begin{table}[t!]
\begin{center}
\begin{tabular}{|l|ccccc|}
\hline \bf test ex.~complexity: & \bf 3 & \bf 4 & \bf 5& \bf 6& \bf 7\\ \hline
right branching & 0.13 & 0.28 & 0.09 & 0.28 & 0.15 \\
left branching &0.82&  0.4 &  0.39 &  0.34 & 0.33 \\
left, no curriculum &\bf 0.89 & 0.28 & 0.33 & 0.23 & 0.32\\
left, steep curriculum & 0.8 & 0.36 & 0.42 & 0.36 & 0.34 \\
left, slow curriculum & 0.77 & \bf 0.64 & \bf 0.47 & \bf 0.45 & \bf 0.38 \\
\hline
\end{tabular}
\end{center}
\caption{\label{acc-table-srn} Accuracy of SRN as a function of the language and input data complexity. Training data in each run includes examples of complexity up to and including $n$, testing data (disjoint from training) contained examples of complexity exactly $n$. Random baseline is 0.25.}
\end{table}

\section{Data hungriness for different architectures}
Tables \ref{recurs-table-lstm}, \ref{recurs-table-gru}, \ref{recurs-table-srn} report the few-shot compositional generalization of different recurrent models. The models were trained on examples of complexity 1 and 2 and some proportion of examples of complexity 3 (i.e.\ minimal recursice examples). The proportion of recursive examples ranged from 0.0 to 0.8. 

\begin{table}[t!]
\begin{center}
\begin{tabular}{|l|ccccc|}
\hline rec.in train & 0.0 & 0.2 & 0.4&  0.6 & 0.8\\ \hline
average accuracy & 0 & .61 & .76 & .89 & \bf .98\\
perfect accuracy & 0 & 0 & 0 & .4 & \bf .9\\
\hline
\end{tabular}
\end{center}
\caption{\label{recurs-table-lstm} Data hungriness for learning recursion by an LSTM at complexity 3, percentage of data complexity 3 included in training data vs.\ test accuracy. The results are based on 10 runs with different random seeds. We report average accuracy as well as the share of runs with perfect accuracy.  Random baseline for accuracy is .25.}
\end{table}

\begin{table}[t!]
\begin{center}
\begin{tabular}{|l|ccccc|}
\hline rec.in train & 0.0 & 0.2 & 0.4&  0.6 & 0.8\\ \hline
average accuracy & 0.69 & .9 & .83 & .98 & \bf 1\\
perfect accuracy & 0.3 & 0 & 0.2 & .9 & \bf 1\\
\hline
\end{tabular}
\end{center}
\caption{\label{recurs-table-gru} Data hungriness for learning recursion by a GRU at complexity 3, percentage of data complexity 3 included in training data vs.\ test accuracy. The results are based on 10 runs with different random seeds. We report average accuracy as well as the share of runs with perfect accuracy.  Random baseline for accuracy is .25.}
\end{table}

\begin{table}[t!]
\begin{center}
\begin{tabular}{|l|ccccc|}
\hline rec.in train & 0.0 & 0.2 & 0.4&  0.6 & 0.8\\ \hline
average accuracy & 0.75 & .66 & .61 & \bf .88 & .78\\
perfect accuracy & 0 & 0 & 0 & \bf .6 & \bf .6\\
\hline
\end{tabular}
\end{center}
\caption{\label{recurs-table-srn} Data hungriness for learning recursion by an SRN at complexity 3, percentage of data complexity 3 included in training data vs.\ test accuracy. The results are based on 10 runs with different random seeds. We report average accuracy as well as the share of runs with perfect accuracy.  Random baseline for accuracy is .25.}
\end{table}

\section{Generalization to unseen lengths}

To illustrate generalization to unseen (high) lengths, we report the performance of different models when trained on data of complexity 1 through 3 and tested on higher complexities (Table \ref{multiple-higher-table}) or when trained on data of complexity 1 through $n$ and tested on data of complexity $n+1$ (Table \ref{next-higher-table}).

\begin{table}[t!]
\begin{center}
\begin{tabular}{|l|cccc|}
\hline \bf test ex.~complexity: & \bf 4 & \bf 5 & \bf 6 & \bf 7\\ \hline
LSTM &  \bf 0.99 & \bf 0.98 & 0.93 & 0.77 \\ \hline
GRU &  0.96 & 0.95 & \bf 0.95 & \bf 0.96 \\ \hline
SRN &  0.76 & 0.53 & 0.67 & 0.77 \\ 
\hline
\end{tabular}
\end{center}
\caption{\label{multiple-higher-table} Generalization of recurrent models from training examples of complexity 1-3 to examples of higher complexity. Accuracy averaged over 10 runs. Random baseline is 0.25.}
\end{table}

\begin{table}[t!]
\begin{center}
\begin{tabular}{|l|cccc|}
\hline \bf test ex.~complexity: & \bf 4 & \bf 5 & \bf 6 & \bf 7\\ \hline
LSTM & \bf 0.99 & 0.95 & 0.8 & 0.7\\ \hline
GRU & 0.96 & \bf 0.996 & \bf 0.98 & \bf 0.94 \\ \hline
SRN & 0.76 & 0.4 & 0.38 & 0.33 \\ 
\hline
\end{tabular}
\end{center}
\caption{\label{next-higher-table} Generalization of recurrent models from training examples of previous levels of complexity to the next level of complexity absent from training data. Accuracy averaged over 10 runs. Random baseline is 0.25.}
\end{table}

\end{document}